\documentclass[conference]{IEEEtran}
\usepackage{amsmath,amssymb,amsfonts}
\usepackage{algorithmic}
\usepackage{graphicx}
\usepackage{textcomp}
\usepackage{hyperref}
\usepackage{xcolor}
\def\BibTeX{{\rm B\kern-.05em{\sc i\kern-.025em b}\kern-.08em
    T\kern-.1667em\lower.7ex\hbox{E}\kern-.125emX}}
\usepackage[utf8]{inputenc}

\title{Evolution of linkages for prototyping of linkage based robots}

\begin{document}

\author{\IEEEauthorblockN{Emma Stensby Norstein}
\IEEEauthorblockA{\textit{Department of Informatics} \\
\textit{University of Oslo}\\
Oslo, Norway}
\and
\IEEEauthorblockN{Kai Olav Ellefsen}
\IEEEauthorblockA{\textit{Department of Informatics} \\
\textit{University of Oslo}\\
Oslo, Norway}
\and
\IEEEauthorblockN{Frank Veenstra}
\IEEEauthorblockA{\textit{Department of Informatics} \\
\textit{University of Oslo}\\
Oslo, Norway}
\and
\IEEEauthorblockN{Tønnes Nygaard}
\IEEEauthorblockA{\textit{Department of Mechanical, Electronic and Chemical Engineering} \\
\textit{Oslo Metropolitan University}\\
Oslo, Norway}
\and
\IEEEauthorblockN{Kyrre Glette}
\IEEEauthorblockA{\textit{RITMO, Department of Informatics} \\
\textit{University of Oslo}\\
Oslo, Norway}}

\maketitle

\begin{abstract}
Prototyping robotic systems is a time consuming process. Computer aided design, however, might speed up the process significantly.
Quality-diversity evolutionary approaches optimise for novelty as well as performance, and can be used to generate a repertoire of diverse designs. This design repertoire could be used as a tool to guide a designer and kick-start the rapid prototyping process. 
This paper explores this idea in the context of mechanical linkage based robots. These robots can be a good test-bed for rapid prototyping, as they can be modified quickly for swift iterations in design.
We compare three evolutionary algorithms for optimising 2D mechanical linkages: 1) a standard evolutionary algorithm, 2) the multi-objective algorithm NSGA-II, and 3) the quality-diversity algorithm MAP-Elites. Some of the evolved linkages are then realized on a physical hexapod robot through a prototyping process, and tested on two different floors.
We find that all the tested approaches, except the standard evolutionary algorithm, are capable of finding mechanical linkages that creates a path similar to a specified desired path. However, the quality-diversity approaches that had the length of the linkage as a behaviour descriptor were the most useful when prototyping. This was due to the quality-diversity approaches having a larger variety of similar designs to choose from, and because the search could be constrained by the behaviour descriptors to make linkages that were viable for construction on our hexapod platform.

\end{abstract}


\section{Introduction}

Robots are becoming more and more common in everyday life, automating tasks both for individuals and companies. As a larger diversity of robots are being designed, tools to aid and speed up the design process are becoming relevant. Designing a robot by hand is a difficult task that requires expert knowledge and is time consuming \cite{nygaard2019experiences}. One way to speed up the process is to offload some of the design to a computer. The engineer could select from a variety of designs presented by a computer, and then adjust the design to the requirements of the robot. This paper is a study of how this variety of initial designs can be generated in the context of linkage based robots.

\begin{figure}[!t]
\normalsize
\centering
\includegraphics[width=200pt, interpolate=false]{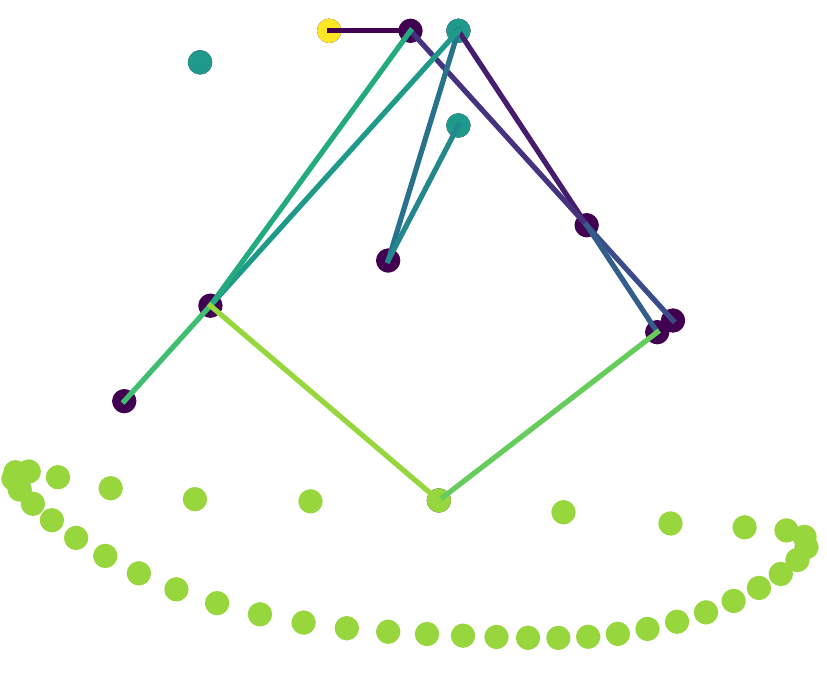}
\caption{A linkage and its path generated by the 2D linkage simulator. The yellow node is the motor, the turquoise nodes are static, and the purple nodes are movable. The green points show the path that the foot of the linkage follows. The color of the beams are arbitrary.}
\label{simulator_example}
\end{figure}

\begin{figure}[!t]
\normalsize
\centering
\includegraphics[width=225pt, interpolate=false]{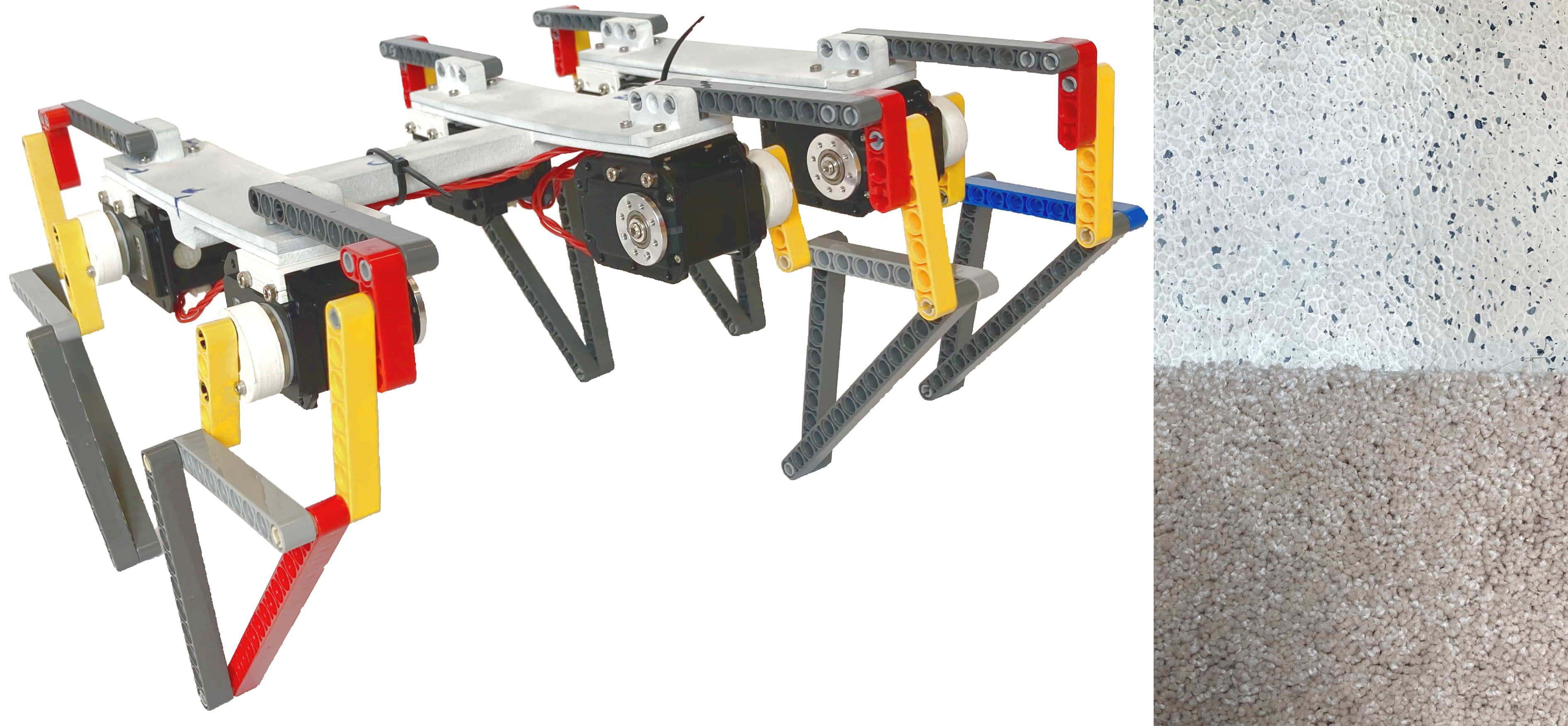}
\caption{The hexapod robot platform used for testing the evolved mechanical linkages, and the two floor textures it is tested on.}
\label{robot}
\vspace{-10pt}
\end{figure}

\begin{figure*}[!t]
\normalsize
\includegraphics[width=\textwidth, interpolate=false]{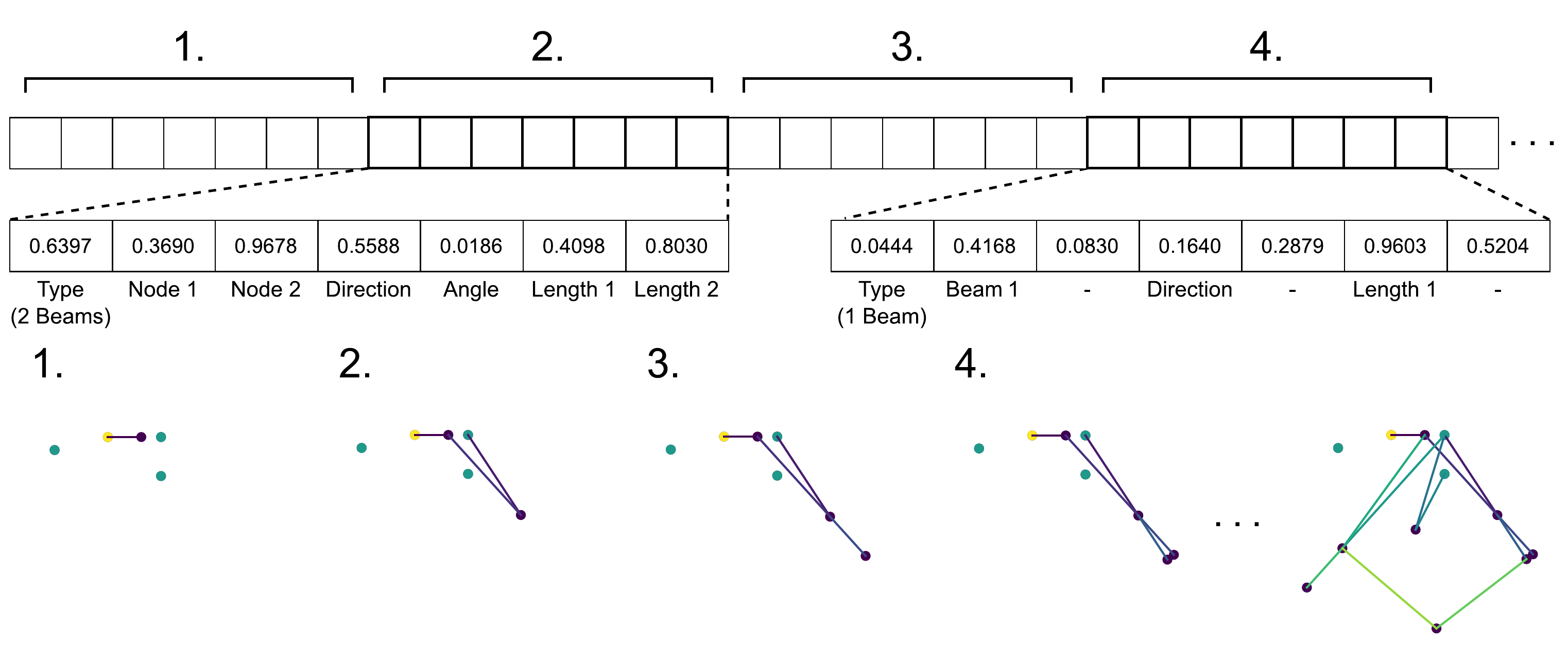}
\vspace{-20pt}
\caption{Decoding of a genome to a linkage. Each section of seven numbers are decoded into one node. The first section is a special section containing the x and y positions of the three static nodes relative to the motor node, and the length of the crank (the beam directly turned by the motor). The first number (type) determines whether the new node is attached by one (below 0.25 as in 4.) or two beams (above 0.25 as in 2.). If the node is attached by one beam the new beam is treated as an extension of an already existing beam. If attached with two beams the new beams can rotate freely. The remaining numbers determine properties of the nodes. The floating point numbers are translated to fit the range of the relevant property. Unlike Type all values in the remaining properties have an equal probability of being chosen. The ranges are as follows: Node: [0, existing nodes), Beam: [0, existing beams), Direction: [0,1], Angle: [0, 2*PI), Length: [Min possible beam length, Max possible beam length]. }
\label{decoding}
\vspace{-10pt}
\end{figure*}

We believe evolutionary algorithms \cite{ea} can be a viable approach to creating initial robot designs. Evolutionary algorithms have a long history in the field of evolutionary robotics for optimising robot morphology and control \cite{erhorizons, HornbyG.S2001Eogd, QDmodular}. A subclass of evolutionary algorithms, that are especially interesting in this context, are quality-diversity algorithms \cite{QDreview}. These algorithms optimise for diversity or novelty along with performance, and often save a repertoire of found solutions \cite{NoveltySearchOriginal}. Quality-diversity algorithms have attracted interest as a method for co-optimising robot morphology and control \cite{QDmodular, zardini2021seeking, NSLC}. 
The quality-diversity algorithm MAP-Elites (Multi-dimensional Archive of Phenotypic Elites) \cite{MAPoriginal}, and extensions such as SAIL \cite{prototypingmapelites}, have previously been used in evolution of modular robots \cite{modularrobots, MAPmodular, MAPmodularAdaptToDamage}, and as a suggestion giver for prototyping in other design tasks \cite{prototypingmapelites, hagg2018prototype}. MAP-Elites can be a powerful tool when prototyping robots. Its diversity promoting repertoires can ensure a wide variety of design suggestions, and can be structured in ways that are easy to navigate. To our knowledge we are the first to combine MAP-Elites as a design suggestion giver with an iterative prototyping process on a physical robot.

To allow for quick design iterations on the physical robot, we choose mechanical linkage based robots as our platform. We find that mechanical linkages is an appropriate test-bed, as new linkages can quickly be created by taking apart and reconnecting beams in different ways, while the chassis of the robot remains the same.
The synthesis of mechanical linkages is a well studied field stretching back over a hundred years \cite{linkagenphard}. Many have used genetic or other optimisation algorithms to create linkages that follow specific paths, often by optimising the length of the beams in a specific linkage configuration \cite{linkagega, fourbar, sixbar}. Mechanical linkages have been used in many industrial applications, and also in robotics for locomotion \cite{designonewalkinglinkage}. Theo Jansen was one of the first to use genetic algorithms to optimise a linkage based walker \cite{jansen2007great}.
Linkage based robots can work with few motors and can therefore be useful for locomotion if a lighter and more energy efficient robot than a regular legged robot is needed. However, linkage based robots are more constrained than regular legged robots. When the robot has been built, the path the legs follow usually cannot be changed without reconstructing the robot, unless self modifying properties have been built into the linkage parts \cite{lengthchangingjansen}. 

We test three evolutionary approaches for generating mechanical linkages to be used on the robot. The three evolutionary algorithms are 1) a standard evolutionary algorithm \cite{ea}, 2) the multi-objective evolutionary algorithm NSGA-II (Non-dominated Sorting Genetic Algorithm II) \cite{nsga2}, and 3) the quality-diversity approach MAP-Elites \cite{MAPoriginal}. We also test two different fitness definitions. The methods are compared with regards to the best fitness found, and on a qualitative analysis of the generated paths. Some linkages are then realised on a hexapod robot through prototyping, while using the repertoires of generated linkages as a design aid. The robots are tested on two floors with different properties, and we attempt to assess how the repertoires can be used to adapt the linkages to the floor textures.

The contributions of this work are twofold: 1) We compare several evolutionary approaches for designing mechanical linkages, in an attempt to find the most efficient method for evolving mechanical leg mechanisms for a hexapod robot. 2) We demonstrate how evolved repertoires of mechanical linkages can be used as part of a prototyping process with a physical linkage based hexapod robot platform.

\section{Methods}

We have implemented a simple 2D simulator in Python for simulating mechanical linkages, which is used to evaluate potential robot designs\footnote{The code and videos of the walking robots can be found at \url{https://github.com/EmmaStensby/linkage-evolution}}. We compare three different evolutionary algorithms for optimising the linkages 1) a standard evolutionary algorithm \cite{ea}, 2) the multi-objective evolutionary algorithm NSGA-II \cite{nsga2}, and 3) the quality-diversity algorithm MAP-Elites \cite{MAPoriginal}. A few solutions from the MAP-Elites approach are tested on a physical robot, when we demonstrate how the repertoires of evolved linkages can be used during robotic system prototyping.

\subsection{2D linkage simulator}
Figure \ref{simulator_example} shows a linkage in the 2D simulator for mechanical linkages. A linkage consists of a crank, three static nodes, as well as several beams connected at free nodes, making up the linkage mechanism. The three static nodes represent where the linkage is connected to the robot, and the crank is the linkage beam that is connected to the motor. The simulator calculates the path each node in the linkage will follow when the crank is rotated one full rotation. The simulator iterates through angles for the crank, and the positions of all the nodes in the linkage are calculated with trigonometry, starting at the static and crank nodes, and moving outwards through the linkage. The lowest node in the linkage is defined as the foot of the robot, as this is the node that will touch the ground. The foot determines the node path returned by the simulator.

\subsection{Linkage encoding}
A linkage is encoded by a vector of floating point numbers within the range 0 to 1. The vector is divided into several sections, the first describing the crank and positions of the static nodes, and each following section describing a node that is added to the linkage. A newly added node will be attached to the linkage by either one or two beams. A node connected with only one beam will always be statically connected to another beam, and will only rotate along with its connected beam to simplify the simulation. If a node is connected by two beams, the beams can rotate freely. This simplification gives two benefits 1) the position of the foot of the linkage is deterministic based on the position of the crank, and will thus always move along the same path with each rotation of the crank, and 2) the position of the foot can be calculated with trigonometry for each crank position, making the simulation of the linkage very efficient. The decoding of a linkage encoding is described in Figure \ref{decoding}.

\subsection{Mutation}
During the evolution the linkages are mutated to create new linkages. The mutation is performed on the encoding representing the linkage, and the linkage is then rebuilt from the new encoding. The encoding is mutated by adding Gaussian noise defined by $N(0,\sigma)$. The noise is always added to the values representing beam length, and added to the rest of the values with a probability of 0.2. $\sigma$ is self adaptive \cite{selfadaptive}. It is encoded along with the genome as a floating point number, and is mutated by adding Gaussian noise defined by $N(0,0.1)$, with a probability of 0.2. After the noise has been added to all values, including $\sigma$, they values are restricted back to the range 0 to 1 using bounce back \cite{nordmoen2021restricting}. The number of nodes in a linkage is constant, but how many of them are involved in the moving mechanism varies.

\begin{figure}[!t]
\normalsize
\centering
\includegraphics[width=225pt, interpolate=false]{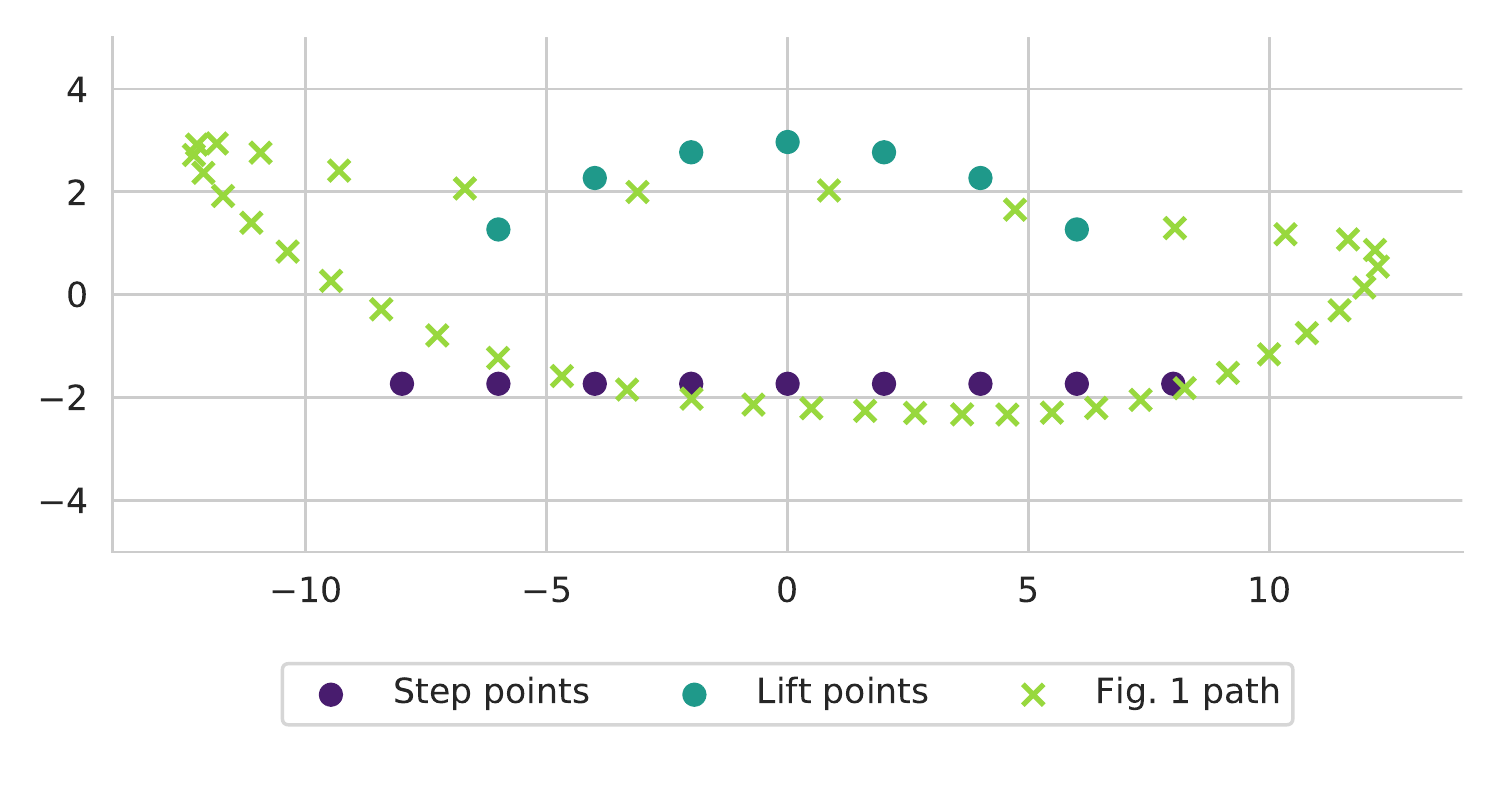}
\caption{The points used to measure the fitness in $F_p$ (Equation \ref{fp}), along with the path from Figure \ref{simulator_example} for reference. $F_p$ for the path shown would be the sum of the distances from each goal point to its closest cross.}
\label{sample_path}
\vspace{-10pt}
\end{figure}

\subsection{Fitness functions}

We compare two fitness functions for all our approaches. For each of the two fitness function there is a separate multiobjective variant that is used by NSGA-II. Both fitnesses are derived from the foot path generated by the simulation. The first fitness function, which we call $F_p$, measures how close the foot path is to following a shape defined by a set of points. The benefit of the first fitness function is that it is easy to define if you know the path you want the leg to follow beforehand. The second, which we call $F_{sl}$, measures step length and leg lift. It is more difficult to design, but is more general.

The points used in the first fitness function are placed in a laying D shape. The shape has a flat bottom to encourage contact between the end of the leg and the ground as the robot takes a step, and a few points above the step line to encourage lifting the leg between the steps. The fitness is calculated as
\begin{equation}
    F_p = -\sum_{i=0}^{n}D_i
    \label{fp}
\end{equation}
where $n$ is the number of points in the set, and $D_i$ is the distances from point $i$ to the closest point in the path. In the multiobjective variant for NSGA-II, the points are divided into two sets, \emph{step points} and \emph{lift points}, and the fitness is measured separately for each of the two. See Figure \ref{sample_path} for the positions and division of the points.

The second fitness function measures the lift and step length of the path. The step length, $F_S$ is measured as the length of the sections of the path that are within a threshold of 5mm from the bottom point in the path in the y-axis direction. The step is only measured on the section of the path where the foot is moving in the same x-axis direction as it is moving in the bottom point of the path, to avoid the foot moving back and forth along the bottom line. The lift, $F_l$ is measured as the maximum y-axis distance from a point on the bottom line to a point above it where the foot is moving in the opposite direction. An angle error, $F_{ae}$ is also added to discourage having the foot of the leg upside down. The angle error is an integer equal to the number of positions where the foot is pointing upwards. These components are combined in $F_{sl}$ as
\begin{equation}
    F_{sl} = - 0.8*F_s - 0.2*F_l + F_{ae}
    \label{fsl}
\end{equation}
The multiobjective variant of $F_{sl}$, used by NSGA-II, has the two objectives $F_{ms} = - F_s + F_{ae}$ and $F_{ml} = - F_l + F_{ae}$.

\begin{figure*}[!t]
\normalsize
\includegraphics[width=\textwidth, interpolate=false]{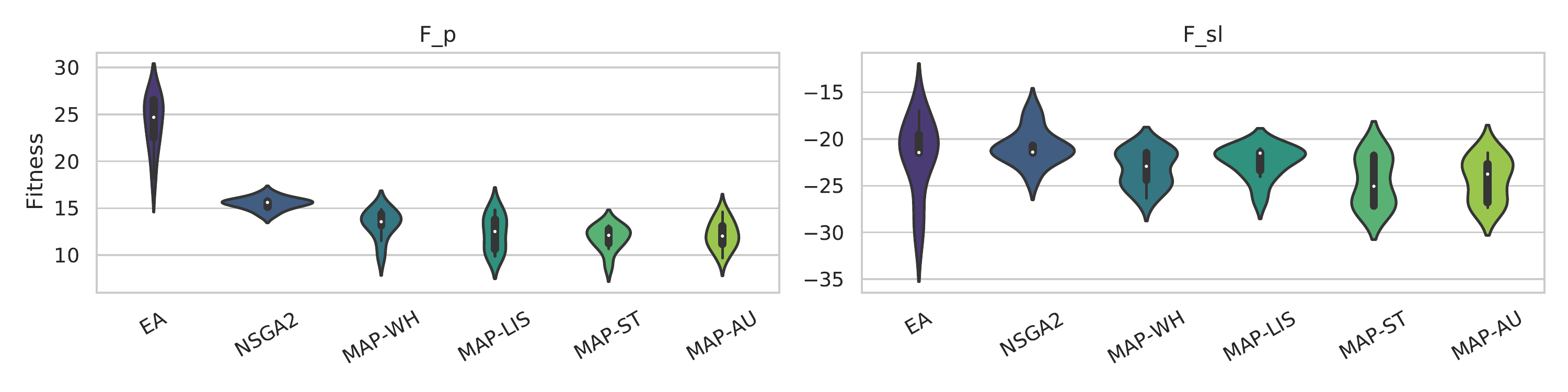}
\caption{Distribution of the best fitness found over the 10 trials of each approach.}
\label{fitness}
\vspace{-10pt}
\end{figure*}

Many linkages will not be solvable for all crank angles. Instead of removing the linkages that cannot turn a full rotation, an error is calculated. The error is an integer equal to the number of crank positions that do not satisfy the linkage constraints. This error is added to \emph{all} fitness functions to discourage non-working linkages.

\subsection{Evolutionary algorithms}
The python library DEAP \cite{deap} was used to implement the evolutionary approaches. All three approaches follow the general structure: 1) \emph{Create children}, 2) \emph{Evaluate linkages} and 3) \emph{Select survivors}, which is repeated until the trial terminates. For all three approaches 5000 individuals were selected from the population to be mutated to create the children. For the standard evolutionary algorithm and NSGA-II this is the entire population, but for MAP-Elites these are selected randomly from the \emph{repertoire} MAP-Elite keeps. The child linkages are then evaluated, and survivors for the next iteration are selected. The standard evolutionary algorithm uses tournament selection with a tournament size of 3, and NSGA-II uses NSGA-II selection \cite{nsga2}. The MAP-Elites selection is described below along with the MAP-Elite repertoires. Parameters were chosen through hand tuning.

\subsection{MAP-Elites}
In MAP-Elites the population is structured as a repertoire of cells, where \emph{behaviour dimensions} decide which cell each solution is placed in. Only one individual is stored in each cell, and the selection of the individual for the cell is elitist.

Four sets of behaviour dimensions are tested for the MAP-Elites approach to get an idea of what properties of the design is most useful to examine when prototyping. The first set, which we call MAP-WH (MAP-Width-Height), has two dimensions based on path shape, the width of the path along the x-axis, and the height of the path along the y-axis.

The second set, which we call MAP-LIS (MAP-Lift-Structure), has two dimensions, one based on linkage structure and one based on path shape. The dimensions are the average length of the beams in the linkage, and the lift of the path. The lift of the path is calculated in the same way as it is for the $F_{sl}$ fitness function.

The third set, which we call MAP-ST (MAP-Structure), has four dimensions based on the structure of the linkage, the average length of the beams, the longest path between a connection to the robot and the foot, the number of the nodes that contribute to the function of the mechanism, and the proportion of moving to stationary nodes.

The last set, which we call MAP-AU (MAP-Aurora), has four dimensions that are automatically defined by an autoencoder \cite{Autoencoder} that is trained on path data throughout the trial, using AURORA (AUtonomous
RObots that Realize their Abilities) \cite{aurora}. As the autoencoder is trained on the paths found it gradually produces a better representation of the search space of possible paths. The autoencoder has 11 hidden layers with respectively 80-64-48-32-16-4-16-32-48-64-80 nodes. The number of layers were chosen by gradually increasing the size of the autoencoder until it could reproduce a few test paths. The values in middle layer with 4 nodes determines the placement of a path in the map. The autoencoder uses the Adagrad \cite{adagrad} optimiser and mean absolute error loss. The autoencoder is solely trained on paths with no error. To initialise the map dimensions the autoencoder is trained for 3000 epochs on the valid paths out of 5000 randomly generated linkages. After this the autoencoder is trained for 1000 epochs every 10 iterations of the evolutionary algorithm, on all valid paths currently in the population.

Both of the maps with two dimensions have a resolution of 100, while the maps with four dimensions have a resolution of 10, giving a total of 10.000 cells for every map.

\subsection{Physical robot}

To be able to quickly test many different linkage configurations, we create the linkages using LEGO Technic bricks. This makes the linkages easily reconfigurable, and the length of the beams in a mechanism can easily be modified by manually switching out a brick. A limitation of using LEGO bricks is that they come in specific sizes, so the evolved beam lengths will need to be approximated by choosing the brick closest in length.

The chassis and spacers for the robot were designed in Fusion 360, and 3D-printed using a HP Jet Fusion 540 and Ultimaker 3 Extended printers. The design consists of several smaller parts that are held together with screws. Because the plastic parts are flexible the robot had to be stiffened with an attached metal rod while walking.
The assembled robot can be seen in Figure \ref{robot}. The actuators used in this design are Dynamixel MX28-AT motors.

When assembling an evolved linkage for the robot, several design choices will have to be made by the builder. The linkage is simulated in 2D with no collisions between beams. When building, the beams that overlap will need to be placed on different layers to avoid collision, while still making the mechanism as stable as possible.

\begin{figure*}[!t]
\normalsize
\includegraphics[width=\textwidth, interpolate=false]{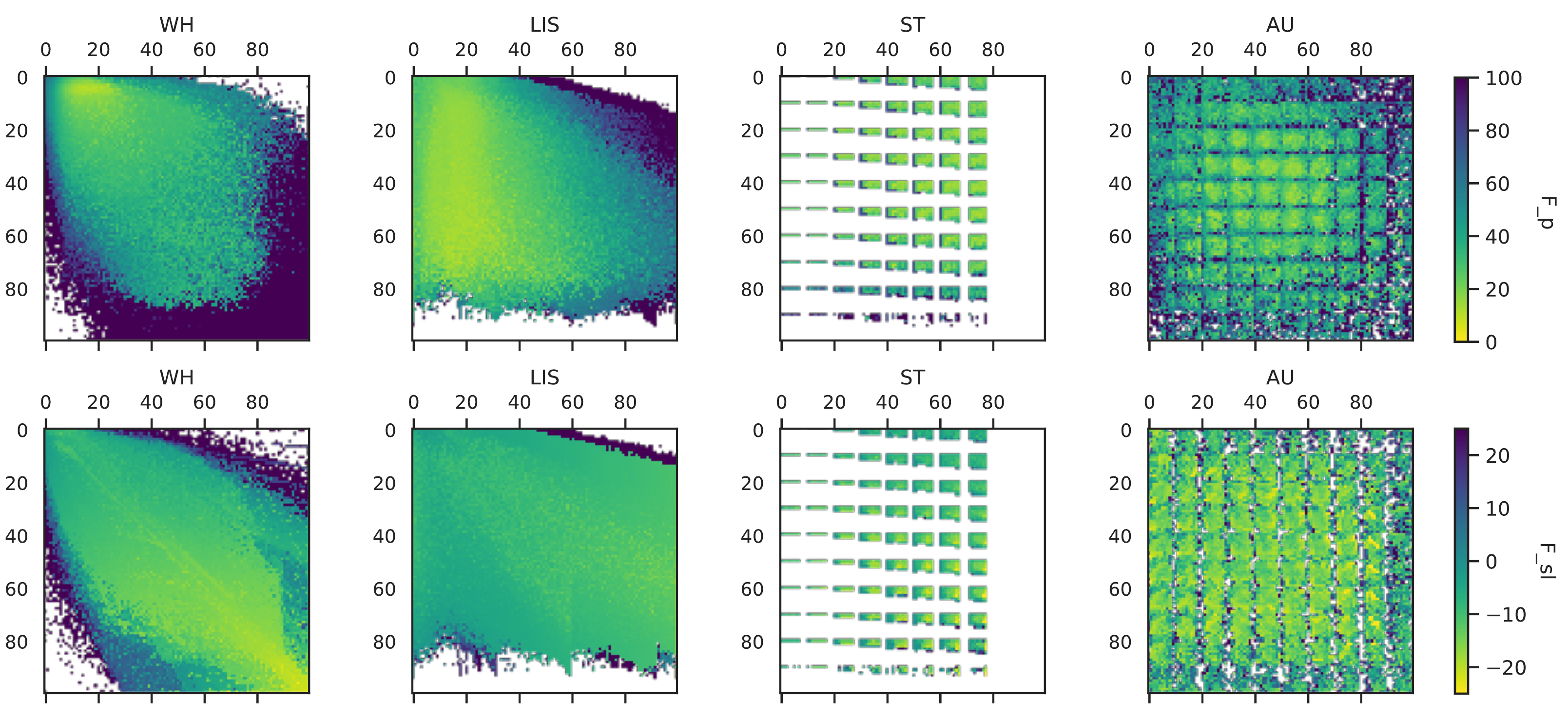}
\caption{The repertoires generated by the MAP-Elites approaches. Each map shows the best fitness found in each cell of the repertoires over all 10 trials of the approach. The color of each cell shows the fitness of the linkage.}
\label{maps}
\vspace{-10pt}
\end{figure*}

\subsection{Prototyping}
The prototyping will consist of building a robot, testing it on two floors with different textures, and then modifying the robot to try to improve its performance. We will do this once for each approach we are comparing. We compare prototyping using repertoires from the four MAP-Elites approaches evolved with the $F_{p}$ fitness function. The first floor is flat and slightly slippery, while the second floor is covered by a thick carpet. We expect the carpet floor to require more lift of the leg as it is not as flat. The floors can be seen in Figure \ref{robot}. The prototyping will follow these stages:
\begin{enumerate}
    \item Build a robot based on the best fitness evolved linkage
    \item Test the robot on both floors
    \item Modify the robot using the evolved repertoires, and try to improve the performance
    \item Test the modified robot on both floors
\end{enumerate}
In stage 3, to find a better design, the maps are downsampled to a size of 5x5 cells, and maps with the paths shown in each cell are created. The original maps contain 10000 cells, so the downsampling ensures the maps are small enough for a human to look through all the presented linkages. We then choose a new linkage from the 5x5 map that has modifications that we believe can improve the robot.

\begin{figure}[!t]
\normalsize
\includegraphics[width=225pt, interpolate=false]{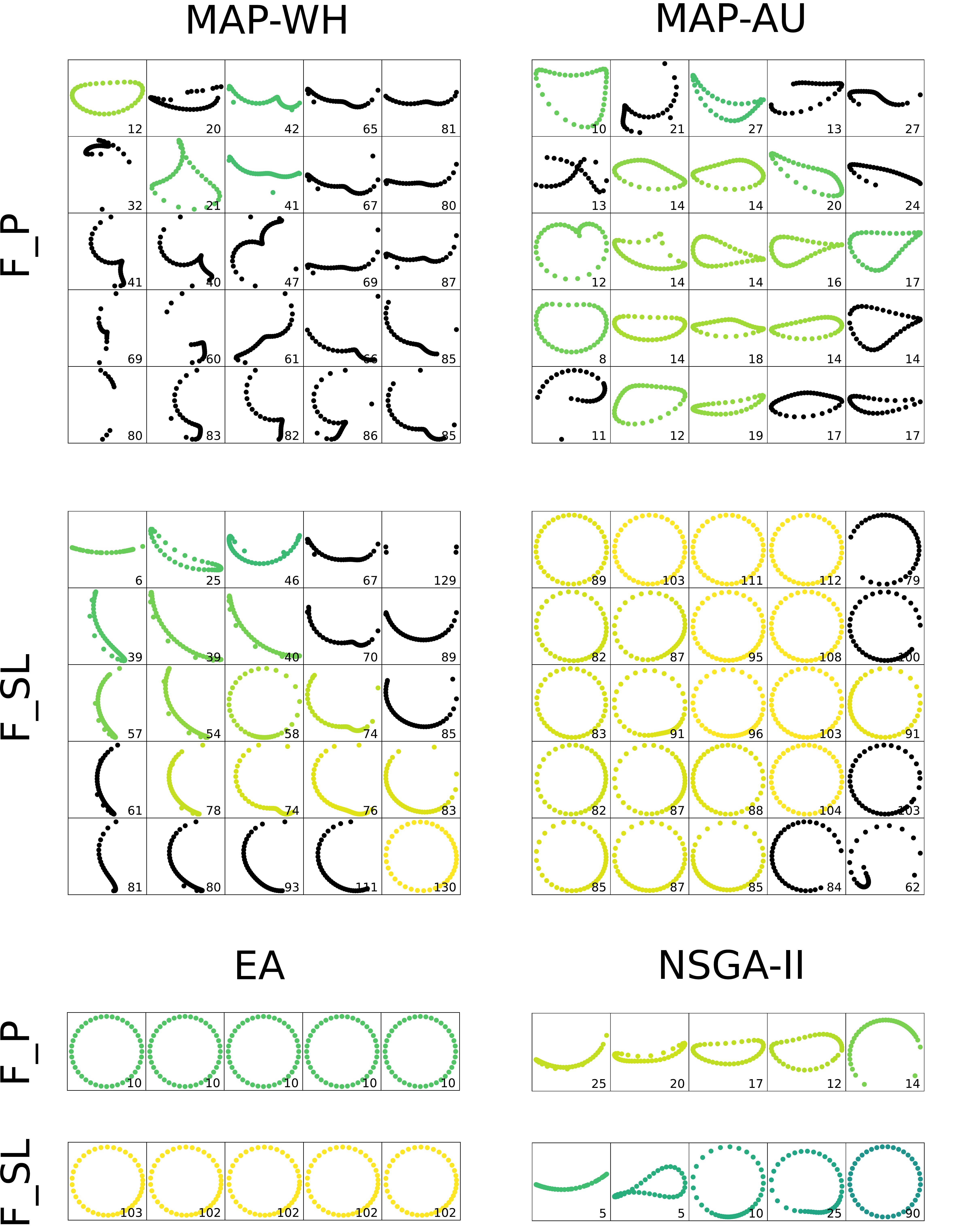}
\caption{Examples of paths generated by some of the approaches. For MAP-Elites the repertoires have been downsampled to a size of 5x5 cells, and the path in each cell is shown. For NSGA-II the paths were sampled from the final pareto front. For the standard evolutionary algorithm the five paths with best fitness from the population are shown.}
\label{paths}
\vspace{-10pt}
\end{figure}

\section{Experiments and Results}

\subsection{2D Simulation}

We run 10 trials of each combination of the evolutionary approaches and fitness functions. Each trial was run for 1 hour, wall clock time, on 16 CPU's. The average number of linkages evaluated per trial was 726750. 

In Figure \ref{fitness} we see the best fitness found by each approach. With the $F_p$ fitness function the map approaches performed better than the evolutionary algorithm, and slightly better than NSGA-II. With $F_{sl}$ all the approaches performed similarly.

The MAP-Elites repertoire maps in Figure \ref{maps} show how the four MAP-Elites approaches explore the search space defined by their dimensions. These maps show the best fitness found in each cell over all 10 trials. The white spaces are areas of the map that were not filled, mostly due to the dimensions in these areas describing properties that are difficult or impossible to combine in a single linkage.

In Figure \ref{paths} we see examples of paths generated in some of the trials. The MAP-WH approach seem to have created more diverse paths in combination with $F_{sl}$, while MAP-AU seem to have created more diverse paths in combination with $F_p$. With the exception of NSGA-II all approaches created linkages that were too large to be successfully realised on the robot when combined with $F_{sl}$.

\subsection{Transfer to reality}

\begin{figure*}[!t]
\normalsize
\label{downsampled}
\includegraphics[width=\textwidth, interpolate=false]{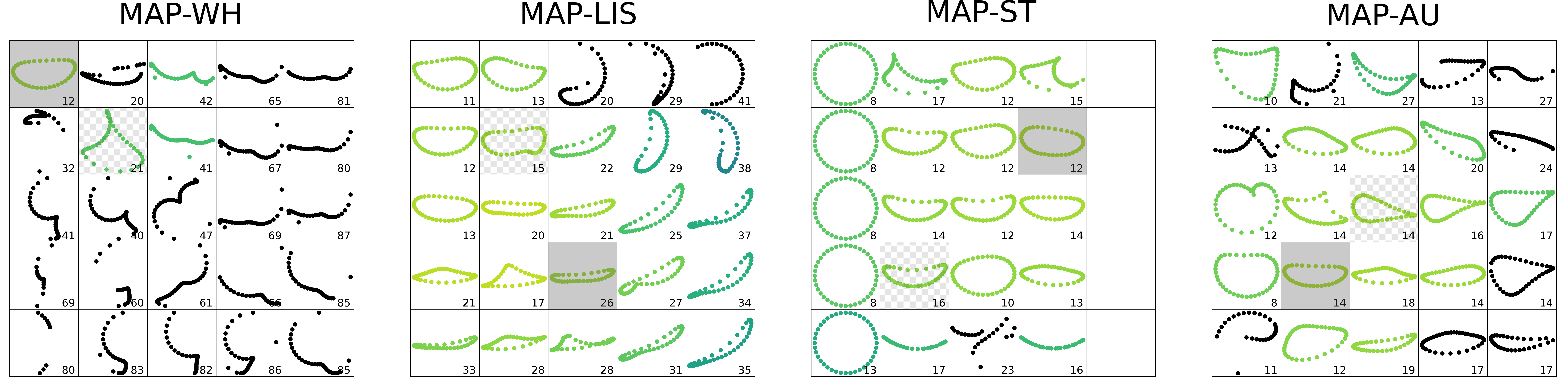}
\caption{Downsampled maps showing the type of linkage paths found in different regions of the maps. The color of the paths indicates their fitness. The number in the bottom right corner of each cell indicates the relative sizes of the paths, as the paths have been scaled to fit their cell. The four grayed out cells show the best fitness paths, and the four checkerboard cells show the paths chosen for the modified robots, together making up the eight linkages tested on the physical robot.}
\label{all_paths}
\end{figure*}

Figure \ref{all_paths} shows downsampled  5x5 maps. The maps are from the four different MAP-Elites approaches using the $F_p$ fitness function. The map cells contain the best fitness path from their region of the original maps. Each of these maps are from a single run of the respective approach, and they were chosen randomly from the 10 trials of each of the four tested approaches. These four maps were used in the prototyping process to choose the modified linkage.

\begin{figure*}[!t]
\normalsize
\includegraphics[width=\textwidth, interpolate=false]{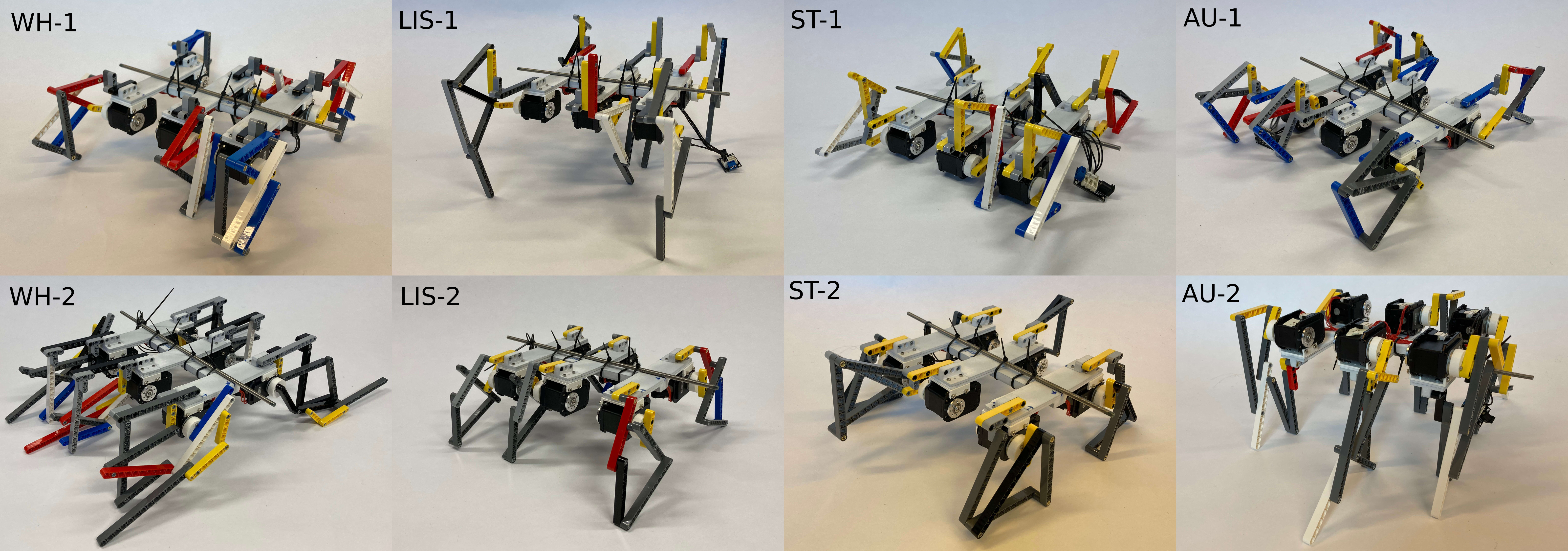}
\caption{The eight robots tested. The top row contains the original robots created based on the best fitness linkage from each map. The bottom row contains the modified robots based on the downsampled repertoires.}
\label{all_robots}
\vspace{-10pt}
\end{figure*}

Images of the eight tested robots are shown in Figure \ref{all_robots}. The second robot on the top row, and the first robot on the second row, were incapable of walking due to their legs being too long and frail to properly support the weight of the robot frame. The rest of the robots were tested five times on each of the two surfaces, with an evaluation time of 10 seconds each. The distance walked in the direction the robot was facing at the beginning of the trial was recorded by hand. The mean distances the six working robots walked on the two floors are summarised in Table \ref{the_table}. %\footnote{Videos of the walking robots can be found at \url{https://github.com/EmmaStensby/linkage-evolution}}

\begin{table}[!t]
\caption{The distances each of the tested robots walked.}
    \label{the_table}
    \centering
    \begin{tabular}{|c|c|c|c|c|}
        \hline
        & \multicolumn{2}{c|}{\textbf{First robot}} & \multicolumn{2}{c|}{\textbf{Modified robot}} \\
        \hline
        \textbf{Approach} & \textbf{Flat floor} & \textbf{Carpet} & \textbf{Flat floor} & \textbf{Carpet}\\
        \hline
        MAP-WH & 73.6 cm & 63.6 cm & \multicolumn{2}{c|}{Breaks (Too long)}\\
        \hline
        MAP-LIS & \multicolumn{2}{c|}{Breaks (Too long)} & 40.4 cm & 48 cm\\
        \hline
        MAP-ST & 39.4 cm & 39.6 cm & 77.6 cm & 39 cm\\
        \hline
        MAP-AU & 61.8 cm & 59.4 cm & 51 cm & 50 cm\\
        \hline
    \end{tabular}
\end{table}

In the first trial, using the MAP-WH repertoires, the original robot created from the best fitness solution walked over 70 cm on the flat floor, in the direction it was facing at the beginning of the trial. It was slightly slower on the carpeted floor. When creating the modified linkage, a larger path with much higher lift was chosen to attempt to increase the speed. However, this linkage was not capable of producing a walking gait. There were not many viable paths to choose between these in the MAP-WH map. This is because the width and height, which are the dimensions in this map, is highly correlated with the fitness, which aims for a path of a certain size. All the viable solutions are therefore likely to be contained within one square on the downsampled map.

In the second trial, using the MAP-LIS repertoires, the original robot had too large legs to support the weight of the robot frame. The modified path was chosen to have a smaller linkage with shorter beams, and a slightly higher step height, while still having many points along the bottom of the path. This robot was not very quick, but was the only one that walked faster on the carpet than the flat floor.

In the third trial, using the MAP-ST repertoires, the original robot walked quite slowly. The original path had quite high lift of the leg, and did not lose any speed on the carpet. The modified linkage was chosen to have a slightly wider path. The modified robot walked quicker on the flat floor, and maintained the same speed as the original on the carpet.

In the fourth and final trial, using the MAP-AU repertoires, the original robot moved quite fast, and similarly on both floors. For the modified linkage we attempted to choose a path that looked a bit flatter on the bottom, with the same amount of leg lift. However, the modified robot moved slightly slower than the original on both floors.

\section{Discussion}

In this study we compared a standard evolutionary algorithm, NSGA-II and MAP-Elites for generating linkage based leg mechanisms. The standard evolutionary algorithm seemed to converge to a local optima with the $F_p$ fitness function, showing that some diversity preservation is beneficial when searching this space. NSGA-II, as expected, produced several linkages representing the tradeoff between its two fitness objectives.

The four MAP-Elites approaches filled the repertoires quite differently, showing the differences between the behavioral dimensions. We qualitatively analyzed the linkages generated by looking at how path shapes were spread throughout downsampled maps.  MAP-AU produced diverse linkages showing that automatically defining the dimensions with an autoencoder is a viable strategy for linkage generation. The benefit of having handcrafted behaviour dimensions is that it can make exploring the repertoires a lot easier, as you know what properties will change when you move through the map. The MAP-WH dimensions based solely on path shape were likely a bit too correlated to the fitness functions, and MAP-ST based solely on linkage structure was difficult to define in a way that filled the entire map. The MAP-LIS which had a combination of path shape and linkage structure as its dimensions seemed the most promising.

Out of the two fitness functions we compared $F_p$ produced the paths that were most viable to recreate on the physical robot. It was intuitive how to control the solutions found using $F_p$. However, to create the set of points for $F_p$ you need to know exactly what foot path you are looking for. $F_{sl}$, which rewards step length and leg lift, could in theory lead to more diverse paths, since it allows more room for how to achieve these properties. In our experiments though, the paths evolved with $F_{sl}$ quickly exploited size to create a big step length, and thus were not easily recreatable on the physical robot. Even though the $F_p$ fitness function is less general than $F_{sl}$, the combination of the MAP-Elites approaches and $F_p$ was capable of producing diverse paths. 

Our study into MAP-Elites as a design tool for linkage based robots provides an initial investigation with eight tested robots. More thorough physical experiments will be needed to generalize our findings.
In addition we likely got better at building the linkages throughout the experiments, which might have affected the results. As the fitness of the modified robots did not improve in any significant way we cannot determine which repertoire produced robots with the highest fitness. However, some repertoires were more user friendly than the others during the prototyping process. The repertoires of MAP-LIS and MAP-AU were the easiest to use as these had a larger diversity of paths to choose from. The MAP-LIS repertoire was especially easy to choose a modified path from, as the length of the linkage could be inferred from the dimensions of the map. Large linkages did not produce a robot capable of locomotion, and with average beam length as a dimension it was possible to compare mechanisms of different sizes in the prototyping stage.

The linkages were simulated in 2D without collisions between beams. When assembling a robot, we needed to choose which layer to place the beams on, and how to connect them, while maintaining a stable structure. This was difficult for some linkages, especially because the mechanisms became frail when the legs were long. The selection of how to connect the beams likely had a substantial effect on the speed of the robots.

In future work it could be interesting to use sturdier materials to create the linkages. This would likely have enabled us to create even faster robots with longer legs, although it would increase the construction time of each robot. To cope with the added construction time, and to make the prototyping experiments fairer, the robots could be built by different people. Another direction that could be explored is to offload even more of the design to the computer by simulating the linkages in 3D on a simulated robot. In this way the assembling of the linkages and the forces between the beams could also be tested before realising a physical robot. This could also increase the fairness of the linkage comparisons, as it would remove human errors in selecting how to assemble the linkage.

\section{Conclusion}
In this paper we compared several evolutionary approaches for designing mechanical linkages, in an attempt to find the most efficient method for evolving mechanical leg mechanisms for a hexapod robot. We then demonstrate how evolved repertoires of mechanical linkages can be used as part of a prototyping process with a physical linkage based hexapod robot platform, by testing eight evolved linkages in the real world.

We found that the quality-diversity and multi-objective approaches were more efficient than the standard evolutionary algorithm at creating a linkage that moves through a set of points, and that the MAP-Elites approach with linkage size as a behavioural dimension was the most useful during the prototyping. We conclude that quality-diversity seems like a promising approach for linkage generation for robotics, and that for prototyping purposes behaviour dimensions reflecting linkage properties that are likely to be changed during the prototyping process are the most useful.

\section*{Acknowledgments}
This work was partially supported by the Research Council of Norway through its Centres of Excellence scheme, project number 262762. The simulations were performed on resources provided by UNINETT Sigma2\textemdash the National Infrastructure for High Performance Computing and Data Storage in Norway.

\bibliographystyle{IEEEtran}
\bibliography{IEEEabrv,main}
\end{document}